\documentclass{article}

\usepackage{arxiv}

\usepackage[utf8]{inputenc} 
\usepackage[T1]{fontenc}    
\usepackage{hyperref}       
\usepackage{url}            
\usepackage{booktabs}       
\usepackage{amsfonts}       
\usepackage{nicefrac}       
\usepackage{microtype}      
\usepackage{lipsum}
\usepackage{multirow}
\usepackage{graphicx}
\usepackage{algorithm}
\usepackage{algorithmic}
\usepackage{amssymb}

\newcommand{\acknowledgments}[1]{
\vspace{6pt}\noindent{\selectfont\textbf{Acknowledgments:} {#1}\par}}

\newcommand{\funding}[1]{
\vspace{6pt}\noindent{\selectfont\textbf{Funding:} {#1}\par}}

\newcommand{\conflictsofinterest}[1]{%
\vspace{6pt}\noindent{\selectfont\textbf{Conflicts of Interest:} {#1}\par}}

\title{Anomaly Detection in Particulate Matter Sensor using Hypothesis Pruning Generative Adversarial Network}

\author{
  YeongHyeon Park \thanks{Correspondence author: yeonghyeon@sk.com}, Won Seok Park, and Yeong Beom Kim \\
  SK Planet Co.,Ltd.\\
  Seongnam, Rep. of Korea\\
  \texttt{yeonghyeon@sk.com, pwswonder@sk.com, and ybkim@sk.com}
}

\begin{document}
\maketitle

\begin{abstract}
World Health Organization (WHO) provides the guideline for managing the Particulate Matter (PM) level because when the PM level is higher, it threats the human health. For managing PM level, the procedure for measuring PM value is needed firstly. We use Tapered Element Oscillating Microbalance (TEOM)-based PM measuring sensors because it shows higher cost-effectiveness than Beta Attenuation Monitor (BAM)-based sensor. However, TEOM-based sensor has higher probability of malfunctioning than BAM-based sensor. In this paper, we call the overall malfunction as an anomaly, and we aim to detect anomalies for the maintenance of PM measuring sensors. We propose a novel architecture for solving the above aim that named as Hypothesis Pruning Generative Adversarial Network (HP-GAN). We experimentally compare the several anomaly detection architectures to certify ours performing better.
\end{abstract}


\section{Introduction}
\label{sec:introduction}

World Health Organization (WHO) recommends to managing the provides Particulate Matter (PM) level with providing their guideline as shown in Table~\ref{table:pm-guideline} because it can infiltrate into the deeper site of humans via respiratory organ \cite{who2006aqg}. For detail, the PM can trigger not only respiratory diseases \cite{faustini2011respiratory} but also cardiovascular disease \cite{du2016aircardio}, lung cancer \cite{hamra2014airlung}, or some other diseases.

For managing PM level with referring Table~\ref{table:pm-guideline}, the process for measuring the air condition should be preceded. In the Republic of Korea, the Air Korea, operated by Korea Environment Corporation, provides measured values of $SO_2$, $CO$, $O_3$, $NO_2$, $PM_{2.5}$ and $PM_{10}$ with unit of one hour. For informing the air pollution to the public, the higher spatial resolution may be more effective than lower. However, the resolution provided by them is relatively low because of the cost of maintaining the high-end PM measuring sensor.

Two types of sensors can be used for PM measurement, each based on Beta Attenuation Monitor (BAM) and Tapered Element Oscillating Microbalance (TEOM) methods \cite{badura2018sensor, bulot2019sensor}. The characteristic of the BAM-based sensor is higher precision of measurement, but it needs high maintenance costs as in the case of Air Korea. On the other hand, the TEOM-based sensor needs a lower cost than the BAM-based sensor. For the above reason, when using a BAM-based sensor, increasing the spatial resolution it is difficult but it can be eased via using a cost-effective TEOM-based sensor.

\begin{table}[h]
    \centering
    \small
    \caption{Air Quality Guideline (AQG) for Particulate Matter (PM) level management from World Health Organization (WHO) \cite{who2006aqg}. The unit of each value is $\mu$g/$m^3$. When the diameter of PM is $2.5\mu$g/$m^3$ it called $PM_{2.5}$, and when it between $2.5\mu$g/$m^3$ and $10\mu$g/$m^3$, called $PM_{10}$.}
    \begin{tabular}{lrr}
        \toprule
            \textbf{} & \textbf{Annual mean} & \textbf{24-hour mean} \\
        \midrule
            $PM_{2.5}$ & 10 & 25 \\
        \midrule
            $PM_{10}$ & 20 & 50 \\
        \bottomrule
    \end{tabular}
    \label{table:pm-guideline}
\end{table}

The two coefficients, one of them is Pearson’s correlation coefficient and the other is a coefficient of determination for 1-hour averages, are used as proof for the cost-effectiveness of TEOM-based sensor. Those coefficients are already measured as 0.91 and 0.81 respectively \cite{badura2018sensor}. 

The TEOM-based sensor can alternate the BAM-based sensor to monitor the PM level. Thus, We have installed TEOM-based sensors relatively densely in several regions as a trial. However, the limitation of the TEOM-based sensor is that it has a more probability malfunctioning in the measuring process than the BAM-based sensor because it more affected by the external environment; we call overall malfunctions as an anomaly. Thus, we propose the novel architecture for anomaly detection to maintain a TEOM-based sensor. Because, if an anomaly detecting solution is provided with the TEOM-based sensor, the efficiency of the PM level monitoring cost can be made more effective.

The organization of this paper is described as follows. In Section~\ref{sec:related_works}, we summarize previous studies that have efforted for anomaly detection. We present the proposed architecture and experimental results in Section~\ref{sec:proposed_architecture} and Section~\ref{sec:experiments} respectively, and conclude the whole content in final section. We only deal with anomaly detection task of functioning sensors in this paper. Thus, the task after anomaly detection, can be categorized to correcting the collected values or maintain the sensor via engineer, will be handled in future study.

\section{Related Works}
\label{sec:related_works}

Several previous studies have conducted the anomaly detection via classification approach with various methods \cite{himmelblau1992classicfication, tellenbach2011classification, eduardo2013svmclassification, jan2017svmclassification, lee2018cnnclassification}. However, there are some problem for using classification method such as difficulty of collecting diverse abnormal cases and labeling cost from collected data to specific category.

For detecting anomaly, the above problems can be eased by regression-based method such as One-Class SVM \cite{steinwart2005classification}, Auto-Encoder (AE) \cite{mayu2014autoencoder, chong2017autoencoder}, or Long-Short Term Memory (LSTM) \cite{malhortra2015lstmtime, chauhan2015ecglstm}. The idea of regression-based method is such simple. Also, the cost of data preparing to use regression-based machine learning or deep learning-based anomaly detection algorithm is not high. Because we need only the normal case (healthy) data for training, and it does not need the categorization process. Moreover, the labeling task is needed for measuring performance quantitatively, but it is quite simple in this case because each sample needs only checked whether normal or not.

Recently, there are several neural network-based anomaly detection models are published \cite{haowen2018vaekpi, thomas2017anogan, donahue2016bigan, samet2018ganomaly, wang2019advae, zhang2019lvead, klar2019conad}. The generative neural network which trained with variational bound can make the user desired data from the random noise \cite{haowen2018vaekpi, kingma2014vae}. However, generating data from noise as same as they do is not essential in anomaly detection. Moreover, because of using variational bound with distribution assumption, the above deep learning architectures may generate the blurred data.

The Generative Adversarial Network (GAN)-based anomaly detection model is published as named as AnoGAN \cite{thomas2017anogan}. The AnoGAN can generate more sharped data than the variational bound-based model. However, it needs to find the most close generated sample from noise with input data for determining abnormal or not. Because of the above procedure, the throughput of AnoGAN is lower than the typical AE-based architecture. 

Some of the recent research, such as BiGAN \cite{donahue2016bigan} and GANomaly \cite{samet2018ganomaly}, eased the limitation of AnoGAN. These models directly generate the sample via input data and use them for anomaly detection. BiGAN \cite{donahue2016bigan} has the limitation of generating high-resolution samples because encoder and decoder share the parameters. However, GANomaly \cite{samet2018ganomaly} eases the above limitation via simply using separated parameters for encoder and decoder respectively.

One other research tried to generate data consistently with avoiding to generate the blurred sample \cite{klar2019conad}. For achieving their purpose, they apply the multiple hypothesis to the last layer of the generator. Their neural network generates multiple samples and the best sample that close to input data among those is selected. However, one of the unsatisfied things is that they still use the variational bound as same as Variational Auto-Encoder (VAE) \cite{kingma2014vae} for training. Thus it can be regarded that has still restricted for generating clear, not blurred, data.

The preprocessing is one of the additional considerations to construct an anomaly detection system for reducing the computational cost. For example, data dimension reduction via preprocessing can be highly reduce the computational cost \cite{serpico2003dimcomp, yan2006dimstream, thangavel2009dimtheory, balzanella2010dimtime, park2018fared}. However, our PM sensor collects the value at every one hour, so the preprocessing is not needed to reduce computational cost such as time.

\section{Proposed Architecture}
\label{sec:proposed_architecture}

In this section, we present the proposed architecture for anomaly detection in the PM sensor. We compare the property of two kinds of convolution method firstly. Then, we present the structure of the neural network that we propose. Finally, we describe the anomaly detection procedure.

\subsection{Reason for Using 1D Convolution}
\label{subsed:reason1d}

When using the image data as the input, the 2 dimensional (2D) convolutional layer can be used for constructing the neural network typically. However, we use the 1 dimensional (1D) convolutional layer because our data is 1D signal data.

\begin{figure}[h]
    \begin{center}
		\includegraphics[width=0.8\linewidth]{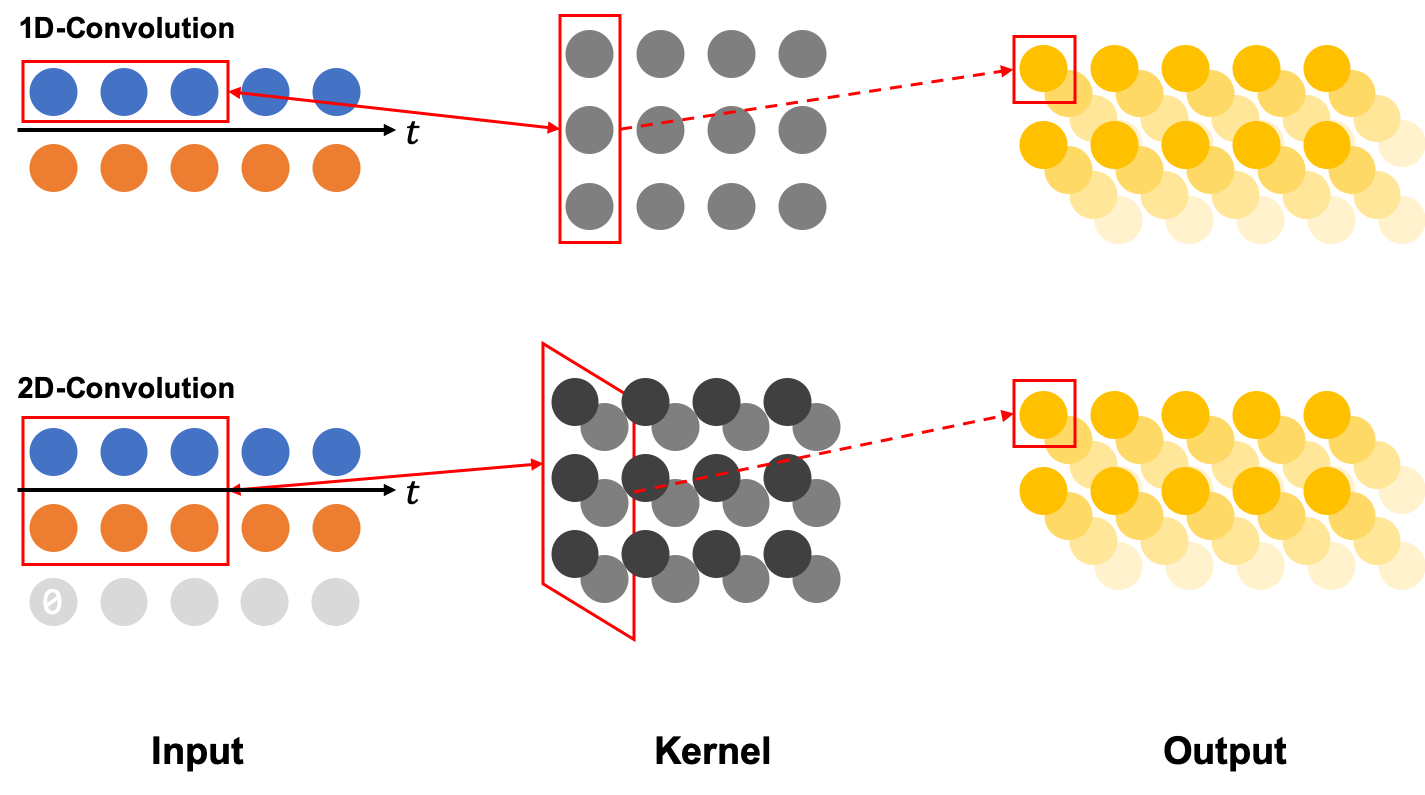}
	\end{center}
	\vspace*{-5mm}
	\caption{The examples of generating the output via 1D and 2D convolution. In 2D convolution case, the zero-padding method is needed for aggregating time information with maintaining feature dimension.}
	\label{fig:conv1d2d}
\end{figure}

The signal data can have the multiple channel. For example, $PM_{2.5}$ and $PM_{10}$ can represent first and second channel respectively. The 2D convolution can be used for processing the multi-channel signal data, but it is not efficient as shown in Figure~\ref{fig:conv1d2d}.

In order to minimize the information loss for each channel, our purpose is only aggregating the time information via a convolutional layer like a recurrent neural network \cite{mikolov2010recurrent}. In the case of 1D convolution, time information can be summarized while maintaining channel information in natural. However, if the neural network constructed with 2D convolution, the channel information of the input data will be reduced unintentionally. For avoiding this problem the zero-padding can be used as shown in Figure~\ref{fig:archtecture}, for maintaining the channel dimension,  but the last channel of the generated output probably has less information than the front channels. Thus, we adopt the 1D convolutional layer for constructing the neural network. 

\subsection{Multiple Hypothesis-based Architecture}
\label{subsed:multiple_hypothesis}

We have already summarized related works in Section~\ref{sec:related_works}. UUsing the variational bound with distribution assumption causes generating blurred the sample. Also, the AnoGAN-like architecture has lower throughput because it needs a matching procedure for finding the most closer generated sample to input data. The helpful thing is multiple hypothesis that can produce more relevant results and can work more robustly \cite{blackman2004multiple, streller2004multiple, rupprecht2017multiple}.

We construct the neural network with reference to the concept of GANomaly as a backbone architecture because it can ease the several mentioned limitations of previous anomaly detection models. We also apply the multiple hypothesis method in the last layer of the generator. Because the multiple hypothesis can utilize to maintain the quality of the output consistently via selecting the best output. We name the procedure of selecting the best output as hypothesis pruning.

\begin{figure*}[h]
    \begin{center}
		\begin{tabular}{cc}
    			\includegraphics[width=0.40\linewidth]{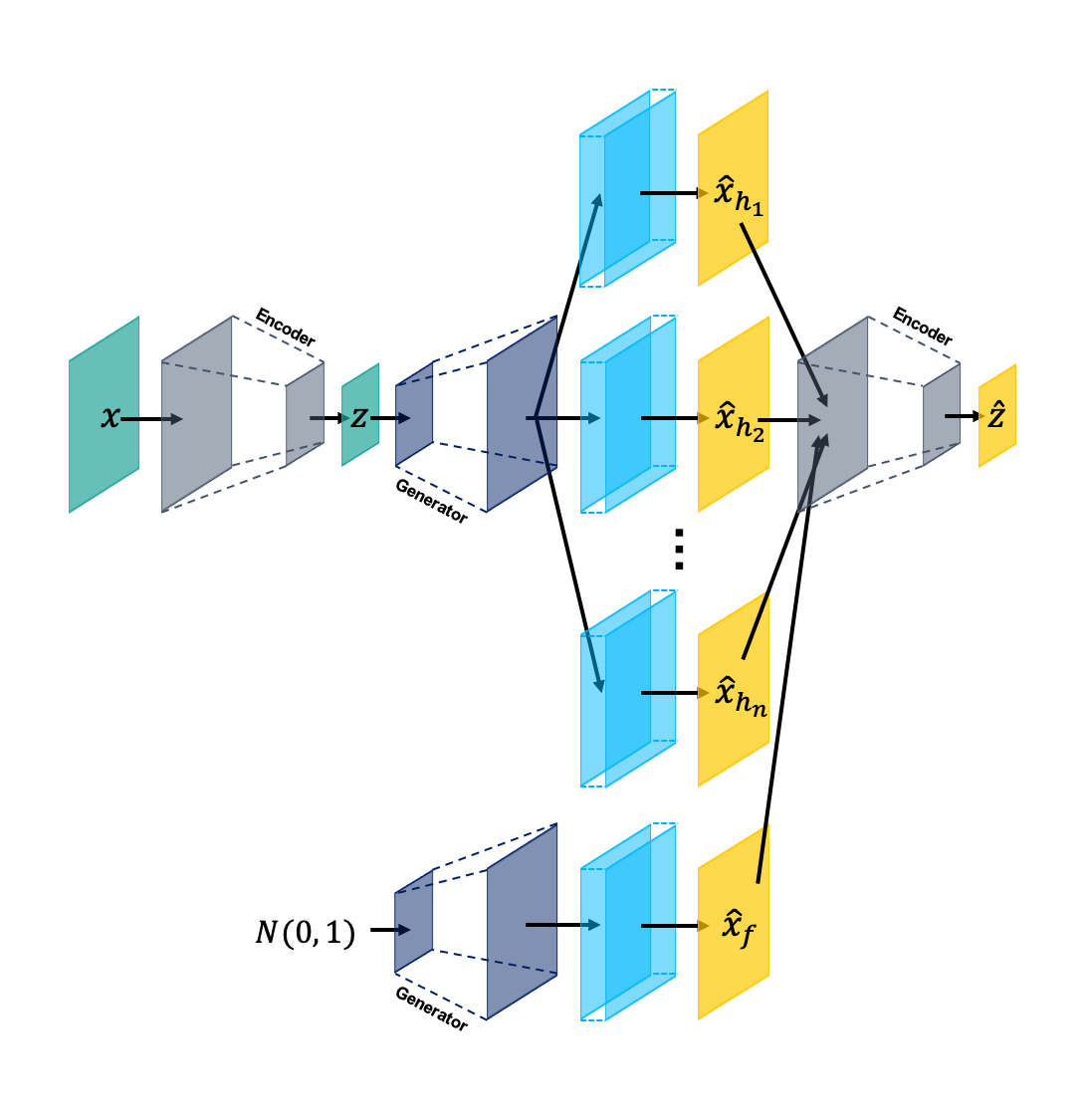} &
    			\includegraphics[width=0.40\linewidth]{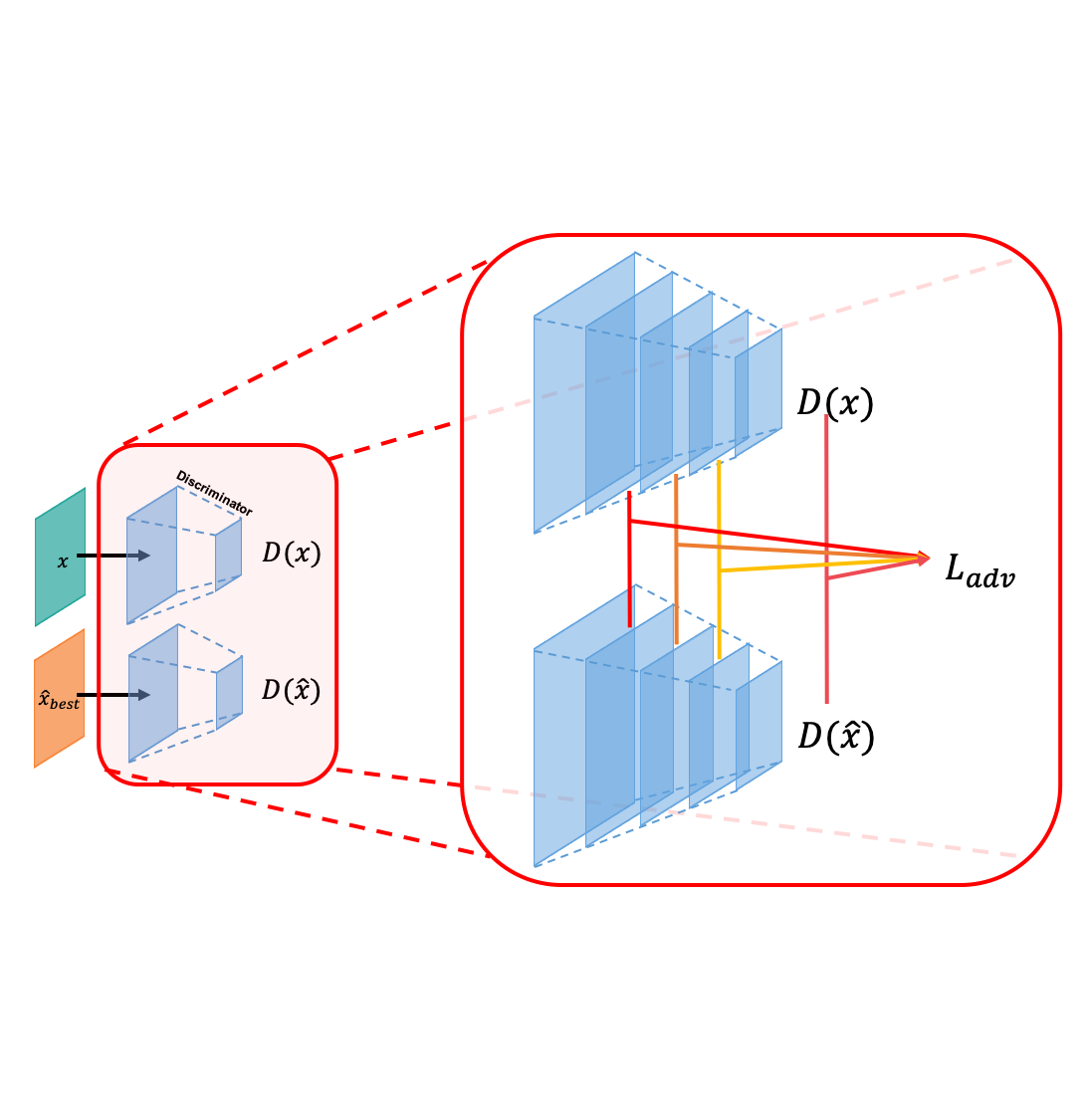}\\
		\end{tabular}
	\end{center}
	\vspace*{-5mm}
	\caption{The architecture of the Hypothesis Pruning GAN (HP-GAN). We use a single encoder, generator, and discriminator but the last layer of the generator is constructed with the multiple hypothesis that colored with cyan. Pruning is conducted after generating the branches by multiple hypothesis.}
	\label{fig:archtecture}
\end{figure*}

The multiple hypothesis generates several samples from the input data directly and generates one additional sample from the random noise. Each generated sample is used as a hypothesis respectively but the samples that are not in the best case will be used as a regularization term. The overall architecture of the neural network that we proposed is shown in Figure~\ref{fig:archtecture}.

The loss functions for training above neural network are shown in Equation~\ref{eq:loss_enc} to Equation~\ref{eq:loss_total}. The symbol $E$, $G$, and $D$ are representing encoder, generator, and discriminator respectively. Also, $x$, $z$, and $f_{l}$ mean input data, encoded latent vector, and feature map of $l$-th layer. The three loss functions $\mathcal{L}_{enc}$, $\mathcal{L}_{gen}$, and $\mathcal{L}_{adv}$ are discripted for encoder, generator, and discriminator respectively; shown in Equation~\ref{eq:loss_enc}, \ref{eq:loss_gen}, and \ref{eq:loss_adv}. Equation~\ref{eq:loss_enc} and \ref{eq:loss_gen}, are constructed with Winner-Take-All (WTA) theory \cite{wolfgang2000wta}. The adversarial losses, from Equation~\ref{eq:loss_adv_noise} to \ref{eq:loss_adv_feature}, are aggregated in Equation~\ref{eq:loss_adv}.

\begin{equation}
    \label{eq:loss_enc}
    \mathcal{L}_{enc} = ||E(x)-E(G(E(x))_{best}||_{2} = ||z-\hat{z}_{best}||_{2}
\end{equation}

\begin{equation}
    \label{eq:loss_gen}
    \mathcal{L}_{gen} = |x-G(E(x))_{best}|_{1} = |x-\hat{x}_{best}|_{1}
\end{equation}

\begin{equation}
    \label{eq:loss_adv_noise}
    \mathcal{L}_{adv_{noise}} = ||D(x)-D(\hat{x}_{noise})||_{2}
\end{equation}

\begin{equation}
    \label{eq:loss_adv_best}
    \mathcal{L}_{adv_{best}} = ||D(x)-D(\hat{x}_{best})||_{2}
\end{equation}

\begin{equation}
    \label{eq:loss_adv_others}
    \mathcal{L}_{adv_{others}} = ||D(x)-D(\hat{x}_{others})||_{2} / \textit{number of others}
\end{equation}

\begin{equation}
    \label{eq:loss_adv_feature}
    \mathcal{L}_{adv_{feature}} = \sum_{l=0}^{L}||f_{l}(x)-f_{l}(\hat{x}_{others})||_{2}
\end{equation}

\begin{equation}
    \label{eq:loss_adv}
    \mathcal{L}_{adv} = \mathcal{L}_{adv_{noise}} + \mathcal{L}_{adv_{best}} + \mathcal{L}_{adv_{others}} + \mathcal{L}_{adv_{feature}}
\end{equation}

For optimizing the parameters at once, we summarize the three losses $\mathcal{L}_{enc}$, $\mathcal{L}_{gen}$, and $\mathcal{L}_{adv}$ with each weighting coefficient $w_{enc}$, $w_{gen}$, and $w_{adv}$. The weighting coefficients are set with values 1, 50, and 1 respectively. These coefficients are referenced from GANomaly \cite{samet2018ganomaly} and they can be modified by the hyperparameter tuning process. We conduct the hyperparameter tuning in Section~\ref{subsec:various_hyperparameter}, but finding the best weighting coefficient is not covered in this paper; we only finding the best kernel size, number of the convolutional block, and learning rate.

\begin{equation}
    \label{eq:loss_total}
    \mathcal{L} = w_{enc}\mathcal{L}_{enc} + w_{gen}\mathcal{L}_{gen} + w_{adv}\mathcal{L}_{adv}
\end{equation}

We train the HP-GAN via Algorithm~\ref{algo:training_algorithm}. We use the Xavier initializer \cite{glorot2010understanding} for initializing the neural network, and use Adam optimizer \cite{kingma2014adam} for optimizing the parameters.

\begin{algorithm}
	\caption{Training algorithm.}
	\hspace*{3.5mm}\textbf{Input:} Measured PM values $x$, and random noise $z_{noise}$\\
	\hspace*{3.5mm}\textbf{Output:} Generated PM values $\hat{X}$ with multiple hypothesis
	\begin{algorithmic} 
		\STATE Initialize parameters of neural network by Xavier initializer \cite{glorot2010understanding}
		\WHILE{the loss has not converged}
            \STATE Get $\hat{X}$ by forward propagation
            \STATE Prune $\hat{X}$ as $\hat{x}_{best}$ or others
            \STATE Compute losses $\mathcal{L}_{enc}$, $\mathcal{L}_{gen}$, and $\mathcal{L}_{adv}$
            \STATE Summarizing losses as a total loss $\mathcal{L}$
            \STATE Update parameters by backward propagation with Adam optimizer \cite{kingma2014adam}
		\ENDWHILE
	\end{algorithmic}
	\label{algo:training_algorithm}
\end{algorithm}

\subsection{Anomaly Detection Method}
\label{subsed:anomaly_detection}

We describe the anomaly detection procedure in this section. The abnormality decision method is simple as shown in Algorithm~\ref{algo:anomaly_detection}. For using the above algorithm, input data must be measured via PM sensor firstly with containing two channels; $PM_{2.5}$ and $PM_{10}$. Then, the generation procedure is conducted via input the data to the neural network. We set the decision boundary $\theta$ using $\mu$ and $\sigma$ of the training data as shown in Equation~\ref{eq:decision}. The $\mu$ and $\sigma$ represent the mean and standard deviation of the mean square error between the input and the best hypothesis of the generated samples. For reference, if the user wants to change the sensitivity or specificity of the anomaly detection, $\theta$ can be adjusted.

\begin{equation}
    \label{eq:decision}
    \theta = \mu + (1.5 * \sigma)
\end{equation}

\begin{algorithm}
	\caption{Abnormality decision algorithm.}
	\hspace*{3.5mm}\textbf{Input:} Measured PM values $x$  \\
	\hspace*{3.5mm}\textbf{Output:} Selected best hypothesis $\hat{x}_{best}$ among the generated samples
	\begin{algorithmic} 
	    \STATE $\theta \leftarrow$ set threshold based on training data
        \IF{$||\hat{x}-\hat{x}_{best}||_{2} \geqslant \theta$}
            \STATE $x$ is abnormal
        \ELSE 
            \STATE $x$ is normal
        \ENDIF
	\end{algorithmic}
	\label{algo:anomaly_detection}
\end{algorithm}

\section{Experiments}
\label{sec:experiments}

In this section, we present our PM dataset and show experimental results for various neural network architectures. For assessing each model, we use Area Under the Receiver Operating characteristics Curve (AUROC) \cite{fawcett2006roc}, Area Under the Precision-Recall Curve (AUPRC) \cite{saito2015prc}, and Mean Square Error (MSE) as the performance indicators.

\subsection{Dataset}
\label{subsed:dataset}

\begin{figure}[h]
    \begin{center}
		\includegraphics[width=0.4\linewidth]{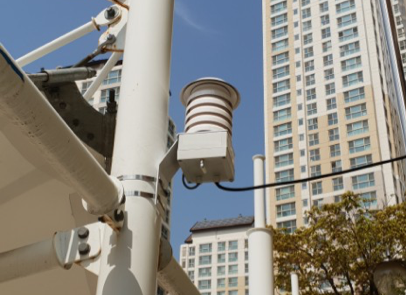}
	\end{center}
	\vspace*{-5mm}
	\caption{TEOM-based PM sensor.}
	\label{fig:sensor}
\end{figure}

We have collected the PM dataset from the 12 locations in the Daegu Metropolitan Jungang Library, Korea. For collecting data, we use the TEOM-based sensor as shown in Figure~\ref{fig:sensor} and the collected dataset is shown in Table~\ref{table:dataset-pm}.

\begin{table}[h]
    \centering
    \small
    \caption{The collected $PM_{2.5}$ and $PM_{10}$ values via TEOM-based sensors. The each sample $x \in \rm I\!R^{24x2}$ represents two values ($PM_{2.5}$ and $PM_{10}$) for 24-hour.}
    \begin{tabular}{cccc}
        \toprule
            \textbf{} & \textbf{Normal} & \textbf{Abnormal} & \textbf{Total} \\
        \midrule
            Number of Sample & 249 & 73 & 322\\
        \bottomrule
    \end{tabular}
    \label{table:dataset-pm}
\end{table}

Each sample $x \in \rm I\!R^{24x2}$ contains the information of $PM_{2.5}$ and $PM_{10}$ for one day with 1-hour unit. The label of normal or abnormal is determined by meteorologists. We use 80\% of the normal for training, and the other normal samples and whole abnormal samples are used for assessing the performance.

\subsection{Experiment with Published Architecture}
\label{subsed:published_architecture}

First of all, we conduct experiments for confirming which architecture among the previous studies is effective for anomaly detection. We adopt five known architectures and reconstruct them using 1D convolutional layer \cite{kingma2014vae, samet2018ganomaly, wang2019advae, zhang2019lvead, klar2019conad}. Each network such as encoder or generator uses three convolutional block, and each convolutional block consists two convolutional layer with elu activation \cite{clevert2015elu}. The max-pooling is applied at the last of each convolutional block. We use two fully connected layers for the rear of the encoder and discriminator. Also, the generator includes two fully connected layers at the front of them. 

\begin{figure}[h]
    \begin{center}
		\begin{tabular}{ccc}
    			\includegraphics[width=0.28\linewidth]{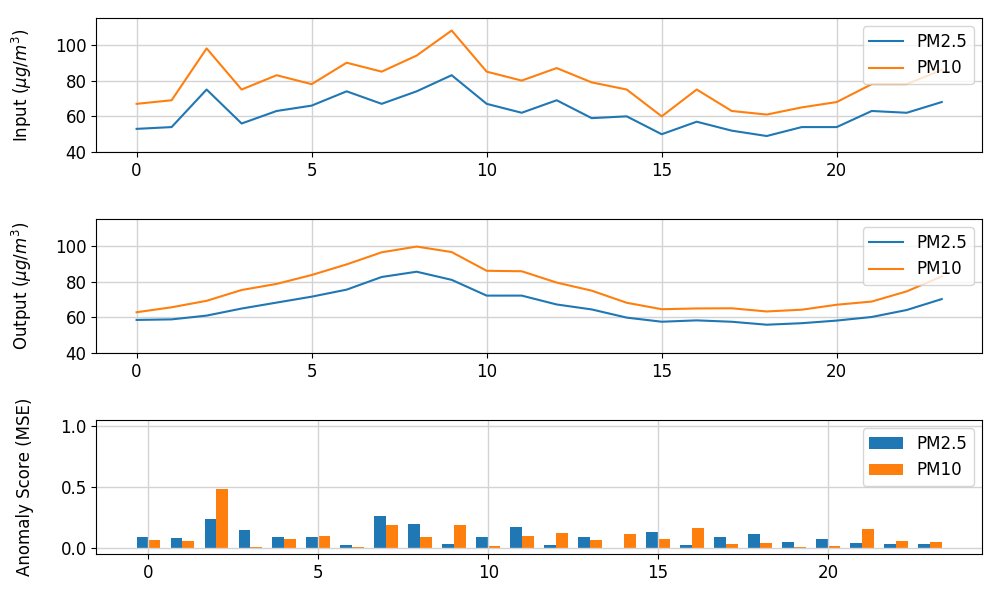} &
    			\includegraphics[width=0.28\linewidth]{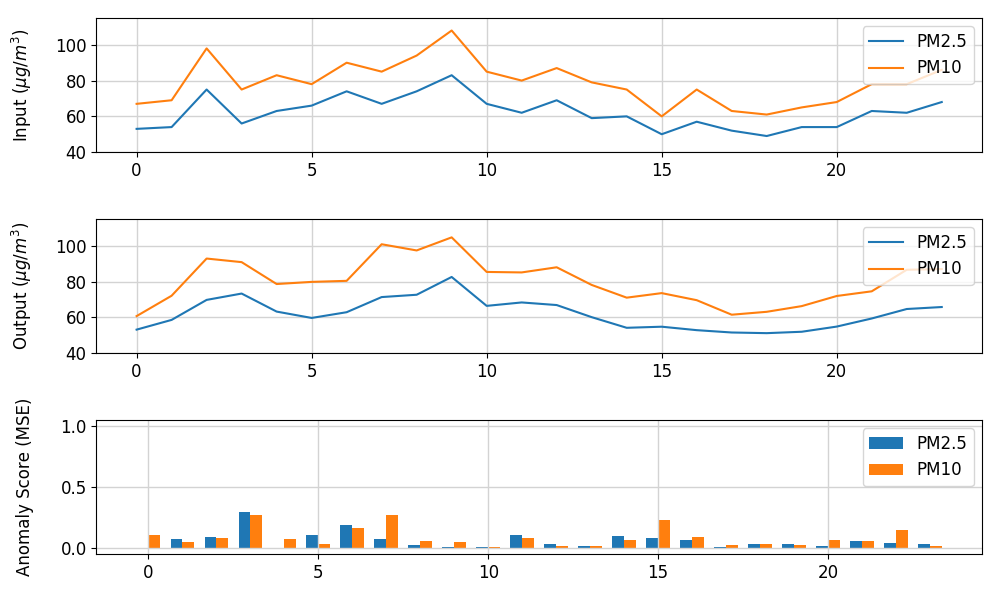} &
    			\includegraphics[width=0.28\linewidth]{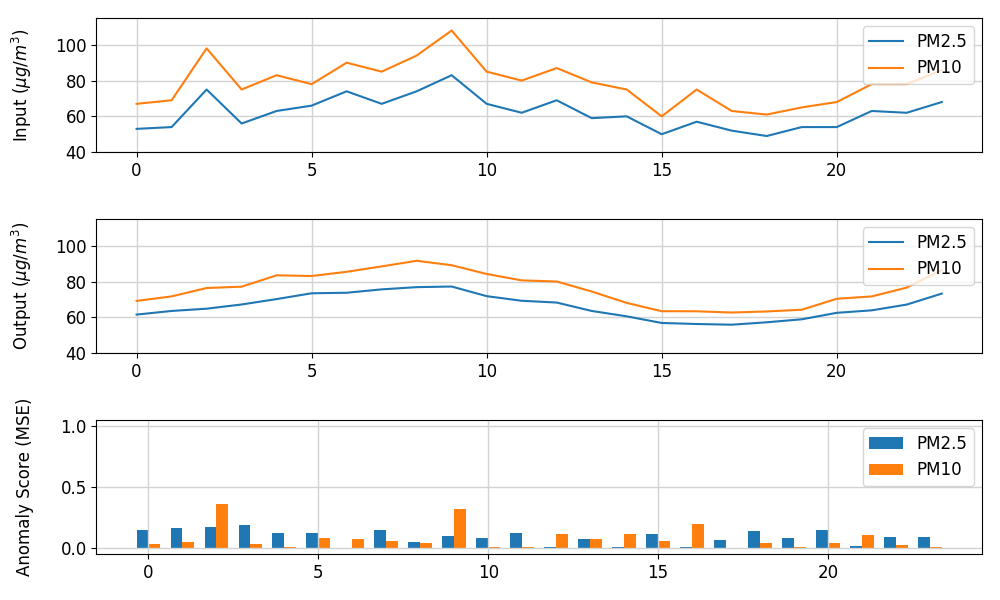} \\
    			VAE & GANomaly & adVAE \\
		    \midrule
    			\includegraphics[width=0.28\linewidth]{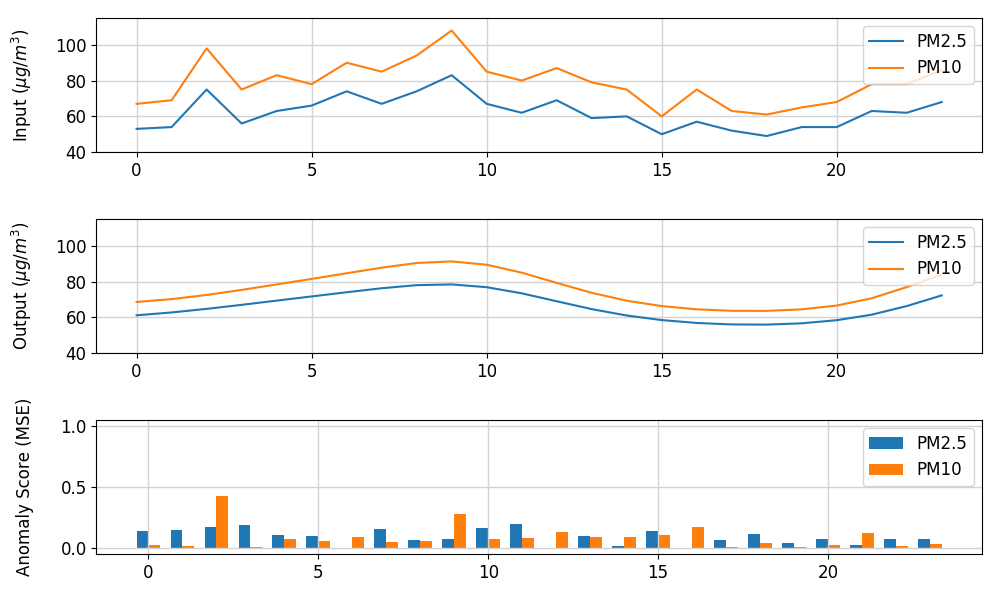} &
    			\includegraphics[width=0.28\linewidth]{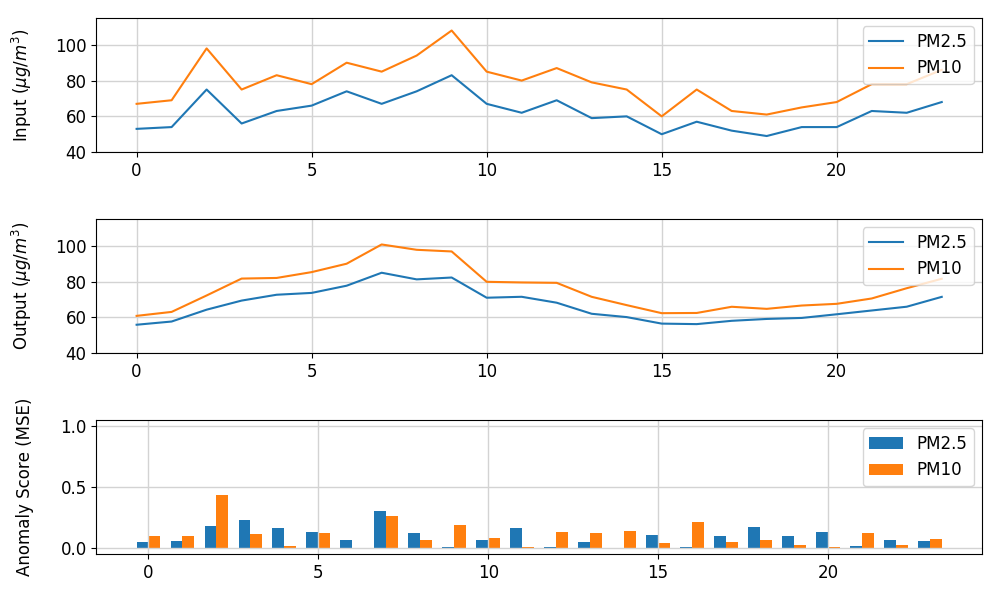} & \\
    			LVEAD & ConAD \\
		\end{tabular}
	\end{center}
	\vspace*{-5mm}
	\caption{The generated sample from the five architectures for qualitative analysis. The x-axis of each subfigure represents an hour of the day and the label of the y-axis is noted in each subfigure.}
	\label{fig:performance-quanti-skdata}
\end{figure}

We measure the performance and summarize them in Table~\ref{table:performance-quali-skdata}. We also present the best-generated samples selected from multiple hypothesis in Figure~\ref{fig:performance-quanti-skdata} for qualitively analysis.

\begin{table*}[h]
    \centering
    \small
    \caption{Measured performance of experiment with published architecture. The purpose of this experiment is confirming which style of the architecture can detect anomaly better.}
    \begin{tabular}{lccccccc}
        \toprule
            \textbf{Aarchitecture} & \textbf{$PM_{2.5}$} & \textbf{$PM_{10}$} 
            & \textbf{AUROC} & \textbf{AUPRC} 
            & \textbf{MSE} & \textbf{Tr-Time} & \textbf{Te-Time} \\
        \midrule
            VAE \cite{kingma2014vae} & O & O & 0.92151 & 0.95629 & 0.01170 & 00:16:43 & 0.41122 \\
                & O & - & 0.91699 & 0.95617 & 0.01424 & \textbf{00:14:26} & 0.30716 \\
                & - & O & 0.92630 & 0.95643 & 0.01772 & 00:17:14 & 0.36180 \\
        \midrule
            GANomaly \cite{samet2018ganomaly} & O & O & \textbf{0.93397} & \textbf{0.96485} & 0.01080 & 00:20:10 & 0.34627 \\
                & O & - & 0.92753 & 0.96209 & 0.00914 & 00:21:43 & 0.31621 \\
                & - & O & 0.92301 & 0.95882 & \textbf{0.01020} & 00:24:21 & \textbf{0.30096} \\
        \midrule
            adVAE \cite{wang2019advae} & O & O & 0.92849 & 0.95704 & 0.01716 & 00:23:42 & 0.40858 \\
                & O & - & 0.92096 & 0.94832 & 0.02646 & 00:19:32 & 0.31841 \\
                & - & O & 0.94233 & 0.95760 & 0.02264 & 00:19:28 & 0.32030 \\
        \midrule
            LVEAD \cite{zhang2019lvead} & O & O & 0.88384 & 0.91724 & 0.01944 & 11:16:55 & 1.03436 \\
                & O & - & 0.85370 & 0.87660 & 0.03764 & 10:37:55 & 1.03876 \\
                & - & O & 0.91575 & 0.94738 & 0.01838 & 11:57:14 & 1.01579 \\
        \midrule
            ConAD \cite{klar2019conad} & O & O & 0.92452 & 0.95638 & 0.02667 & 00:20:22 & 0.34142 \\
                & O & - & 0.92342 & 0.95606 & 0.01524 & 00:30:05 & 0.31843 \\
                & - & O & 0.91616 & 0.94154 & 0.01799 & 00:30:02 & 0.30392 \\
        \bottomrule
    \end{tabular}
    \label{table:performance-quali-skdata}
\end{table*}

Time consumption for training and test procedure of four architectures other than LVEAD are similar to each other but LVEAD needs much longer time. Thus, LVEAD is not efficient in the time consumption viewpoint. 

The GANomaly shows the highest AUROC and AUPRC. Also, it has the lowest MSE among the above five architectures, so if who wants to use the known neural network without developing novel architecture GANomaly can be recommended. 

For analyzing qualitative results, GANomaly generates the most similar sample to the input. The ConAD and VAE follow after GANomaly. The common method of VAE and ConAD is variational bound. The above method can be regarded as a reason for generating smoothed output. Thus, we conduct an ablation study to find and confirm the cause of performance degradation in the next section.

\subsection{Ablation Experiment}
\label{subsed:ablation_experiment}

We construct HP-GAN by referring to previous studies. HP-GAN, the novel unproven architecture, needs to confirm that it can work better than other architectures. We compose the experiments to verify the ability of HP-GAN with our dataset. The several kinds of architectures other than HP-GAN are constructed for ablation experiments. The ablation experiment is used to find the cause of the performance impediment \cite{sajid2015ablation, richard2019ablation}. 

The six ablation architectures are constructed that contain the GANomaly and HP-GAN. Latent vector Matching (LM), Variational Bound (VB), and Multiple Hypothesis (MH) are used as the conditions for construct these architectures. The purpose of using LM is minimizing the Euclidean distance between latent vector $z$ (from $x$) and $\hat{z}$ (from $\hat{x}$). The VB is used for minimizing the Kullback–Leibler divergence between latent vector $z$ and normal distribution. The MH, last condition, is used to generate consistent output based on (WTA) theory.

The mini-batch size, number of the epoch, the dimension of the latent vector $z$, and the learning rate are used equally for all architectures. The above hyperparameters are set as 32, 1000, and 0.0001 respectively. The experimental result is summarized in Table~\ref{table:performance-ablation-skdata}. 

\begin{table*}[h]
    \centering
    \small
    \caption{The performance of the six ablation (combination) architectures. We use three condition Latent vector Matching (LM), Variational Bound (VB), and Multiple Hypothesis (MH) for ablation study.}
    \begin{tabular}{lcccccccc}
        \toprule
            \textbf{Architecture} & \textbf{LM} & \textbf{VB} & \textbf{MH} & \textbf{$PM_{2.5}$} 
            & \textbf{$PM_{10}$} & \textbf{AUROC} & \textbf{AUPRC} & \textbf{MSE} \\
        \midrule
            \multirow{3}{*}{\begin{tabular}[x]{@{}l@{}}Ablation-1\\(LM-GAN; GANomaly)\end{tabular}}
                & \multirow{3}{*}{O} & \multirow{3}{*}{-} & \multirow{3}{*}{-}
                & O & O & 0.93397 & 0.96485 & 0.01080 \\
                & & & & O & - & 0.92753 & 0.96209 & 0.00914 \\
                & & & & - & O & 0.92301 & 0.95882 & \textbf{0.01020} \\
        \midrule
            \multirow{3}{*}{\begin{tabular}[x]{@{}l@{}}Ablation-2\\(VB-GAN)\end{tabular}}
                & \multirow{3}{*}{-} & \multirow{3}{*}{O} & \multirow{3}{*}{-}
                & O & O  & 0.90548 & 0.93130 & 0.02148 \\
                & & & & O & - & 0.94192 & 0.96536 & 0.01924 \\
                & & & & - & O & 0.86767 & 0.91166 & 0.04300 \\
        \midrule
            \multirow{3}{*}{\begin{tabular}[x]{@{}l@{}}Ablation-3\\(LMVB-GAN)\end{tabular}}
                & \multirow{3}{*}{O} & \multirow{3}{*}{O} & \multirow{3}{*}{-}
                & O & O  & 0.89521 & 0.93441 & 0.02510 \\
                & & & & O & - & 0.86096 & 0.90572 & 0.04022 \\
                & & & & - & O & 0.89205 & 0.93045 & 0.02974 \\
        \midrule
            \multirow{3}{*}{\begin{tabular}[x]{@{}l@{}}Ablation-4\\(LMMH-GAN; Ours)\end{tabular}}
                & \multirow{3}{*}{O} & \multirow{3}{*}{-} & \multirow{3}{*}{O}
                & O & O  & 0.91699 & 0.95303 & 0.02550 \\
                & & & & O & - & 0.93219 & 0.96312 & 0.01534 \\
                & & & & - & O & 0.92616 & 0.95424 & 0.01230 \\
        \midrule
            \multirow{3}{*}{\begin{tabular}[x]{@{}l@{}}Ablation-5\\(VBMH-GAN)\end{tabular}}
                & \multirow{3}{*}{-} & \multirow{3}{*}{O} & \multirow{3}{*}{O}
                & O & O  & \textbf{0.94644} & \textbf{0.96462} & 0.03578 \\
                & & & & O & - & 0.93027 & 0.96153 & 0.01486 \\
                & & & & - & O & 0.90068 & 0.94807 & 0.01404 \\
        \midrule
            \multirow{3}{*}{\begin{tabular}[x]{@{}l@{}}Ablation-6\\(LMVBMH-GAN)\end{tabular}}
                & \multirow{3}{*}{O} & \multirow{3}{*}{O} & \multirow{3}{*}{O}
                & O & O  & 0.92603 & 0.94032 & 0.03019 \\
                & & & & O & - & 0.93123 & 0.96179 & 0.01531 \\
                & & & & - & O & 0.92712 & 0.95730 & 0.01922 \\
        \bottomrule
    \end{tabular}
    \label{table:performance-ablation-skdata}
\end{table*}

\begin{figure}[h]
    \begin{center}
		\begin{tabular}{ccc}
    			\includegraphics[width=0.28\linewidth]{figures/pm-publicmodel-ganomaly} &
    			\includegraphics[width=0.28\linewidth]{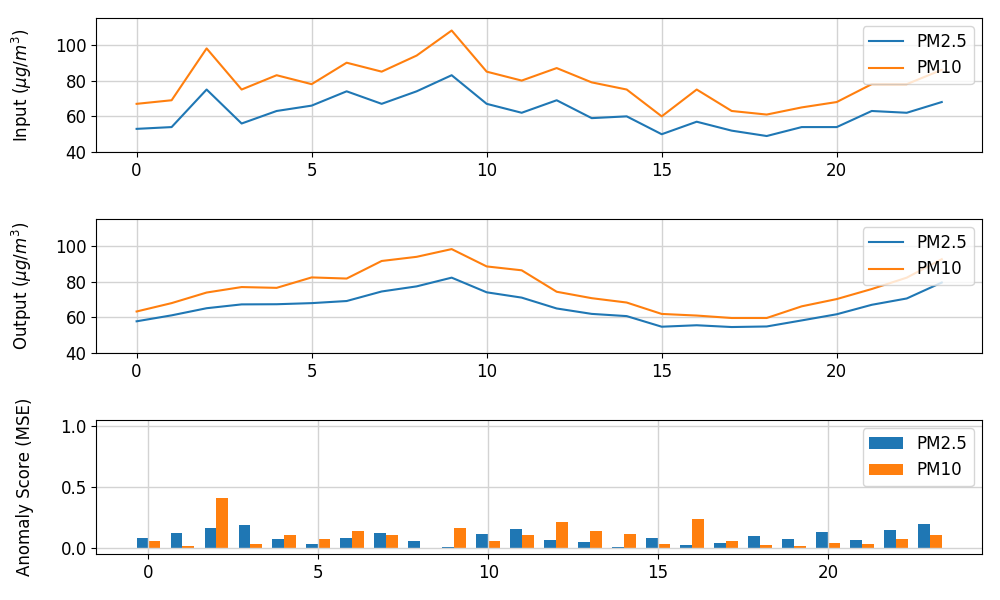} &
    			\includegraphics[width=0.28\linewidth]{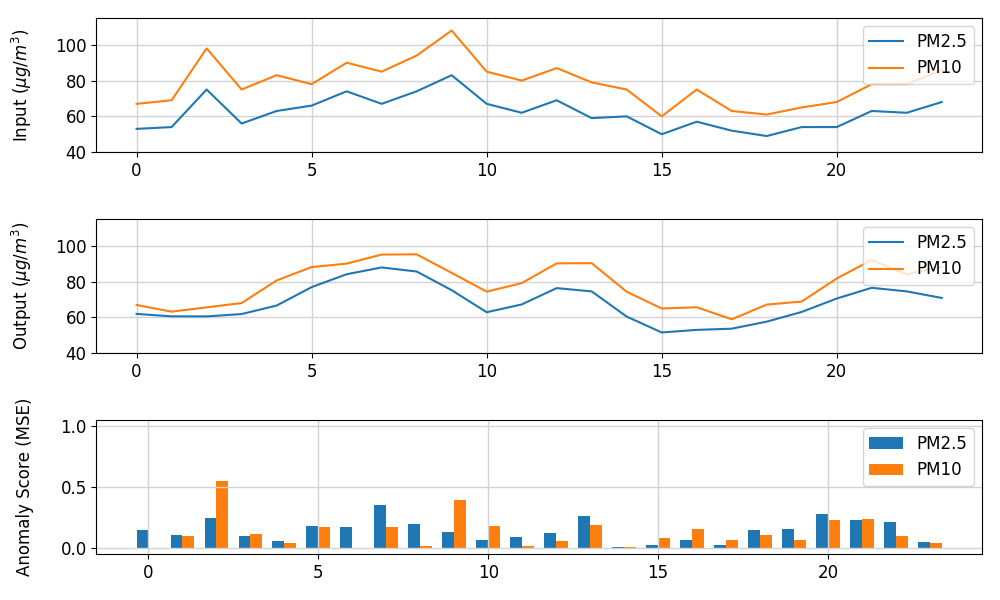} \\
    			LM-GAN (GANomaly) & VB-GAN & LMVB-GAN \\
		    \midrule
    			\includegraphics[width=0.28\linewidth]{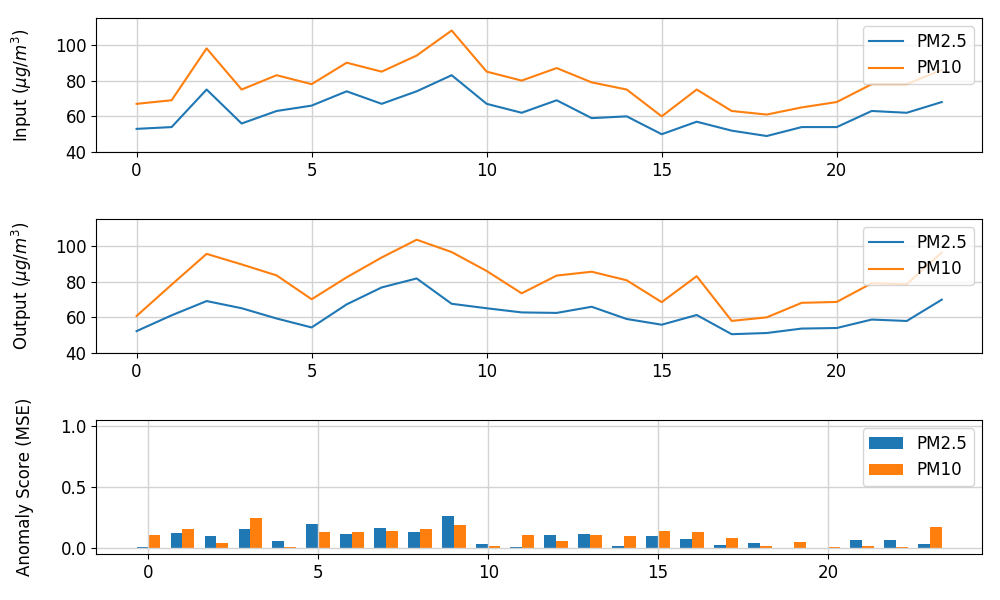} &
    			\includegraphics[width=0.28\linewidth]{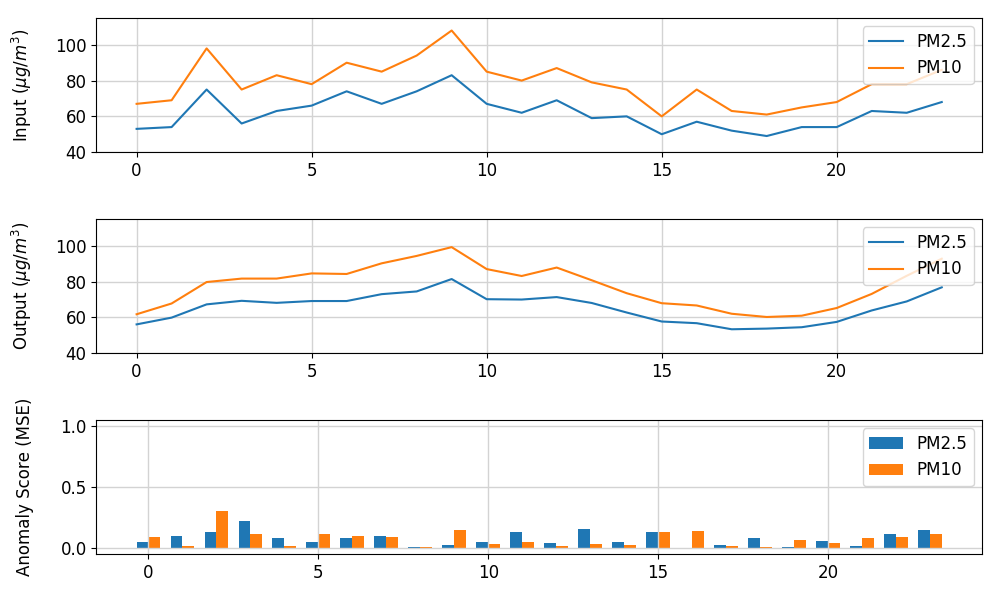} &
    			\includegraphics[width=0.28\linewidth]{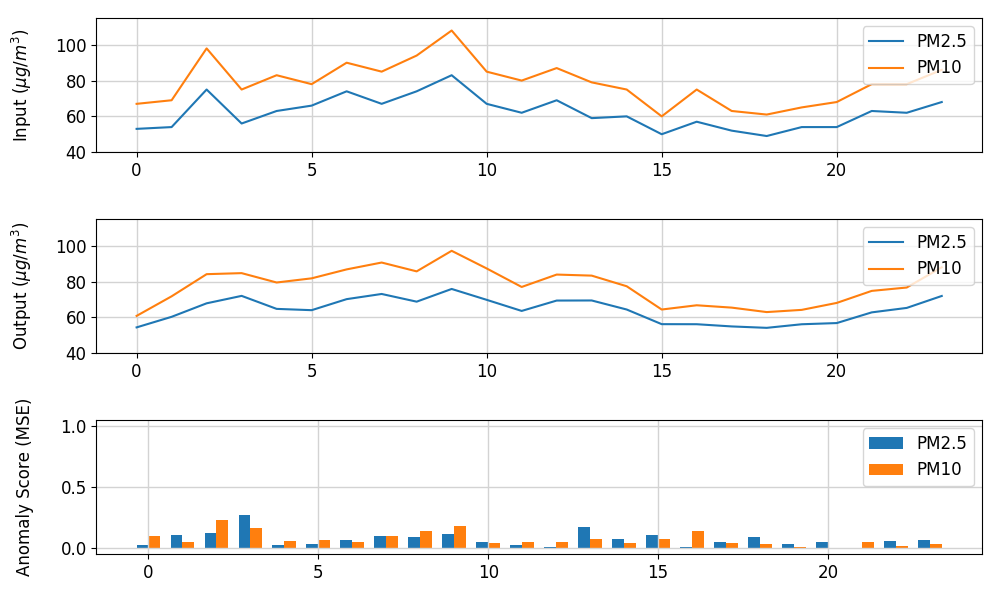} \\
    			LMMH-GAN (Ours) & VBMH-GAN & LMVBMH-GAN \\
		\end{tabular}
	\end{center}
	\vspace*{-5mm}
	\caption{The generated sample from the six ablation architectures for qualitative analysis.}
	\label{fig:performance-ablation-skdata}
\end{figure}

The Table~\ref{table:performance-ablation-skdata} quantitatively shows that VBMH-GAN has better AUROC, AUPRC than the other. However, as shown in Figure~\ref{fig:performance-ablation-skdata}, the qualitative result, the generated sample via LM-GAN (GANomaly) and LMMH-GAN (Ours; HP-GAN) show a better result than VBMH-GAN. Moreover, other VB-based architectures commonly show the blurred result.

In this experiment, we confirm that the quantitative result represents that the VBMH-GAN is the best architecture, but the qualitative result of them is not good. Thus, we need to conduct more experiment for verifying the above architectures using varied hyperparameters, because each model may need the specific hyperparameter for performing much better.

\subsection{Experiment with Various Hyperparameter}
\label{subsec:various_hyperparameter}

In this section, we assess four of neural networks with various hyperparameter set. For the experiment, we use the VAE as a baseline architecture because it is the first VB-based architecture. We also use GANomaly, based on LM, as one other baseline architecture. The other two architectures are our HP-GAN (LMMH-GAN) and VBMH-GAN.

We use the kernel size, number of convolutional blocks, and learning rate as the hyperparameter. Each convolutional block consists two convolutional layer with elu activation and one max pooling layer same as commented in Section~\ref{subsed:published_architecture}. We compose the 108 kinds of the hyperparameter set for experiment via combining three hyperparameters as shown in Table~\ref{table:hyperparameter_set}.

\begin{table}[h]
    \centering
    \small
    \caption{The hyperparameters for the experiment. We use the grid search method for hyperparameter tuning \cite{chicco2017ten}. The number of convolutional blocks is abbreviated as \# of conv-block.}
    \begin{center}
		\begin{tabular}{ll}
		    \toprule
		        \textbf{Hyperparameter} & \textbf{Values} \\ 
			\midrule
    			Kernel size & 3, 5, 7, 9, 11, and 13\\ 
                \# of conv-block & 2, 3, and 4  \\
                Learning rate & 5e-3, 1e-3, 5e-4, 1e-4, 5e-5, and 1e-5 \\ 
            \bottomrule
		\end{tabular}
	\end{center}
	\label{table:hyperparameter_set}
\end{table}

When the case of large kernel sizes such as 13, the time axis dimension of the feature vector can be smaller than the kernel of the last layer. However, the amount of the feature information and balance between feature vectors is not changed by the above situation, so we also use the large size kernels.

\begin{figure}[h]
    \begin{center}
		\begin{tabular}{ccc}
    			\includegraphics[width=0.30\linewidth]{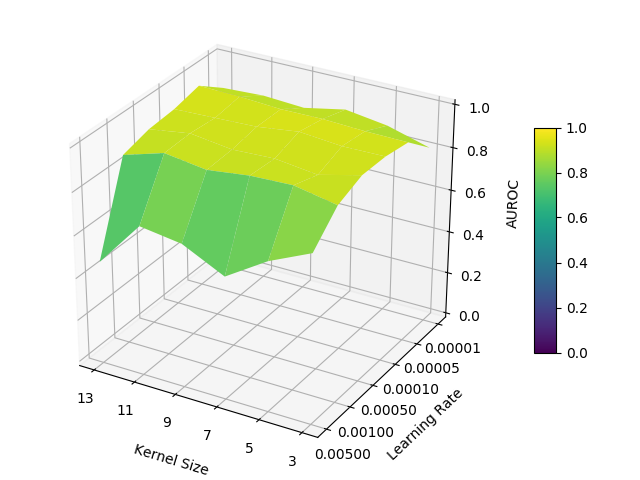} &
    			\includegraphics[width=0.30\linewidth]{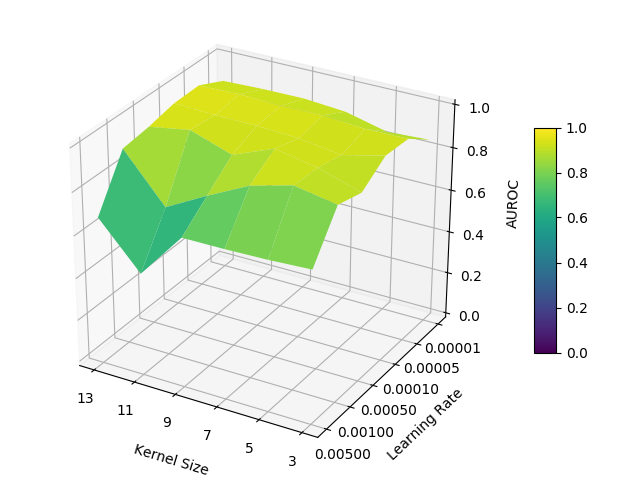} &
    			\includegraphics[width=0.30\linewidth]{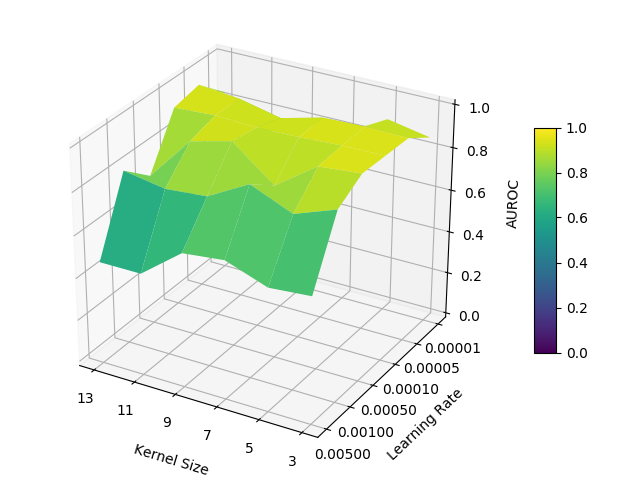} \\
			\midrule
    			\includegraphics[width=0.30\linewidth]{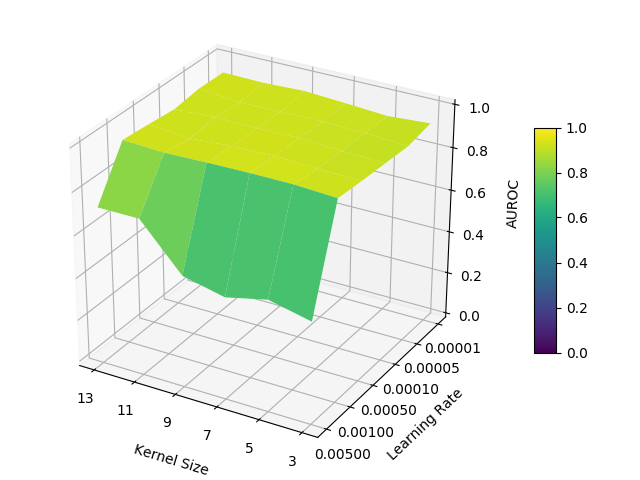} &
    			\includegraphics[width=0.30\linewidth]{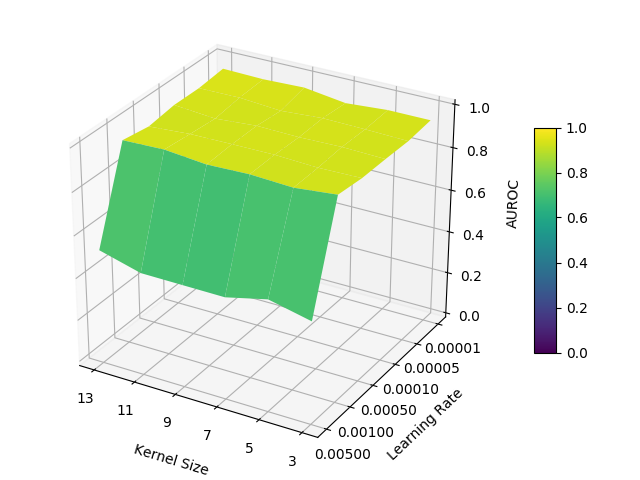} &
    			\includegraphics[width=0.30\linewidth]{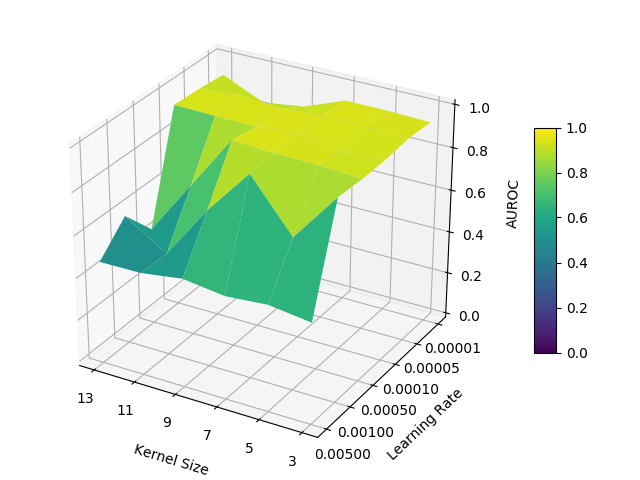} \\
			\midrule
    			\includegraphics[width=0.30\linewidth]{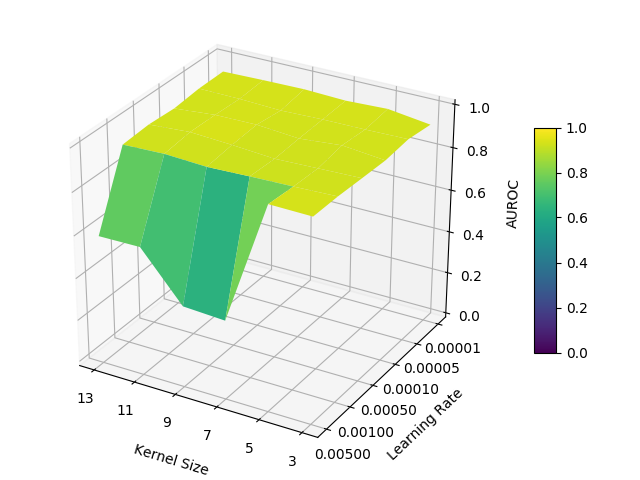} &
    			\includegraphics[width=0.30\linewidth]{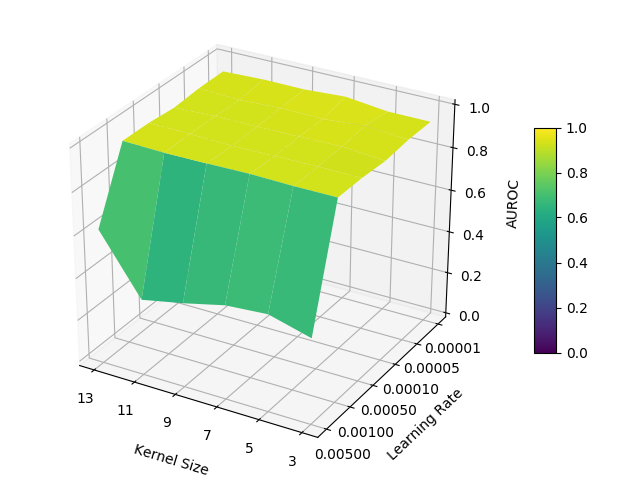} &
    			\includegraphics[width=0.30\linewidth]{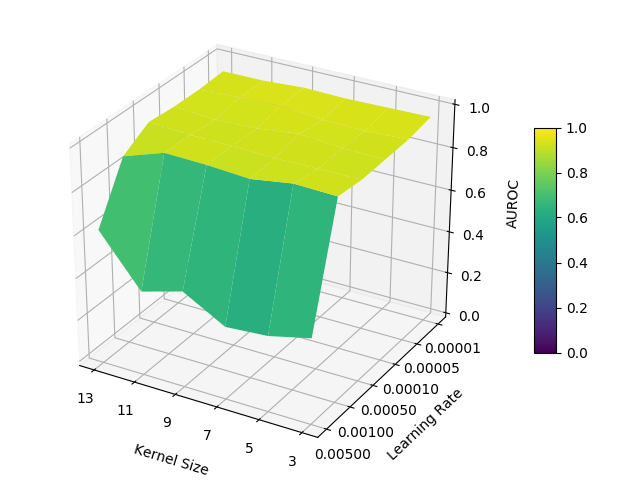} \\
			\midrule
    			\includegraphics[width=0.30\linewidth]{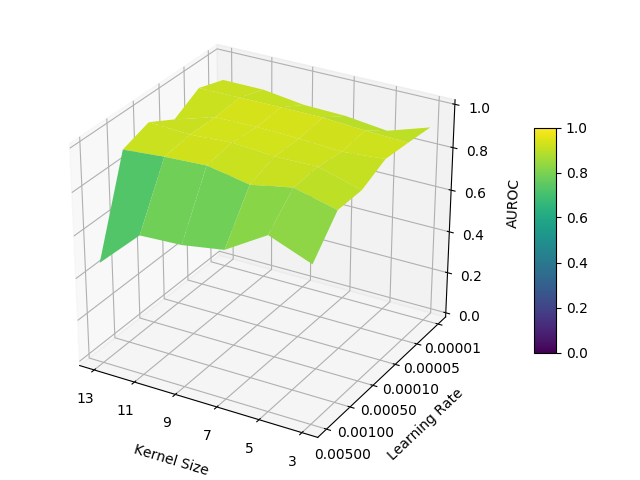} &
    			\includegraphics[width=0.30\linewidth]{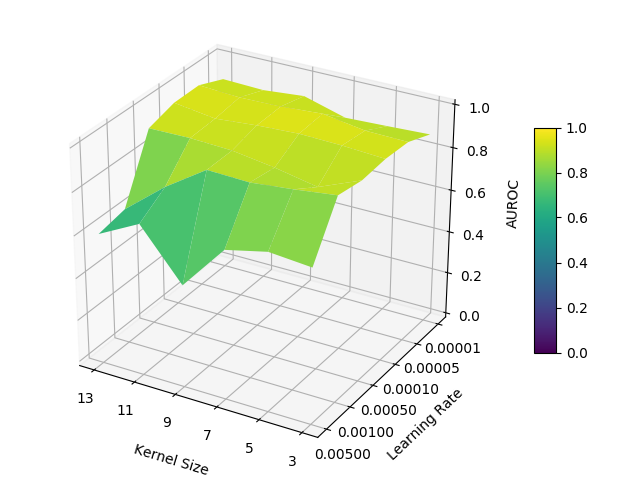} &
    			\includegraphics[width=0.30\linewidth]{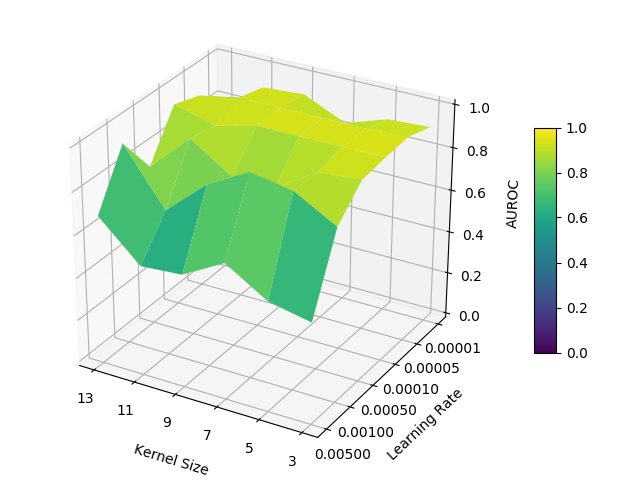} \\
		\end{tabular}
	\end{center}
	\vspace*{-5mm}
	\caption{The surface of AUROC with various hyperparameters. The AUROC of VAE, GANomaly, and HP-GAN (LMMH-GAN), and VBMH-GAN are shown in each row sequentially. Each column shows the result of two, three, and four convolutional blocks from the left to right.}
	\label{fig:hyperparameter_auroc_sk}
\end{figure}

\begin{figure}[h]
    \begin{center}
		\begin{tabular}{ccc}
    			\includegraphics[width=0.30\linewidth]{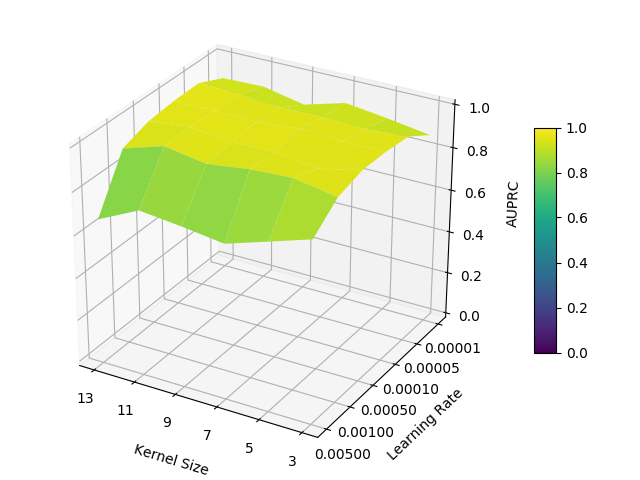} &
    			\includegraphics[width=0.30\linewidth]{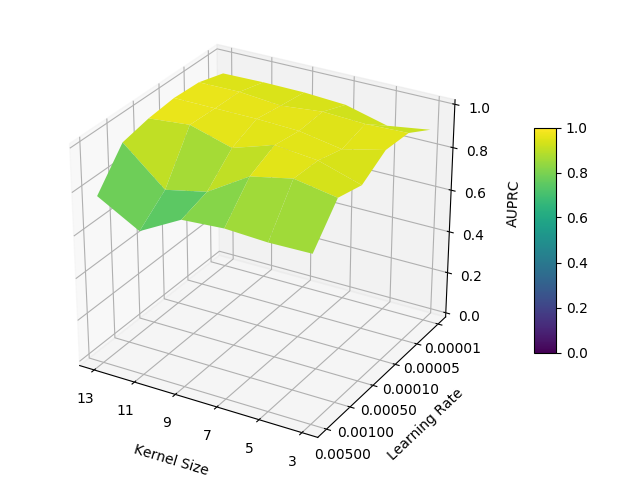} &
    			\includegraphics[width=0.30\linewidth]{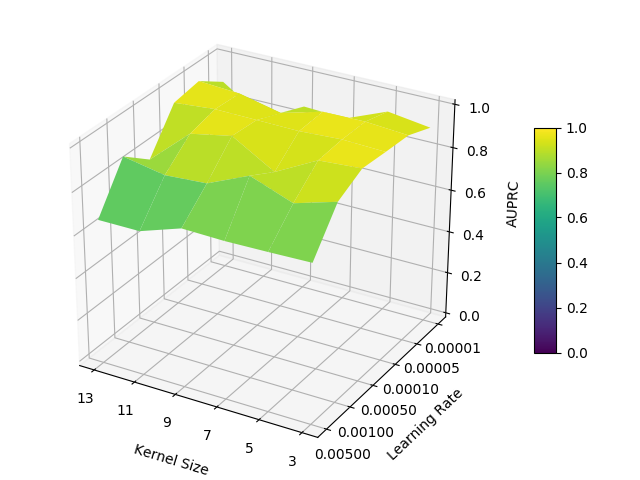} \\
			\midrule
    			\includegraphics[width=0.30\linewidth]{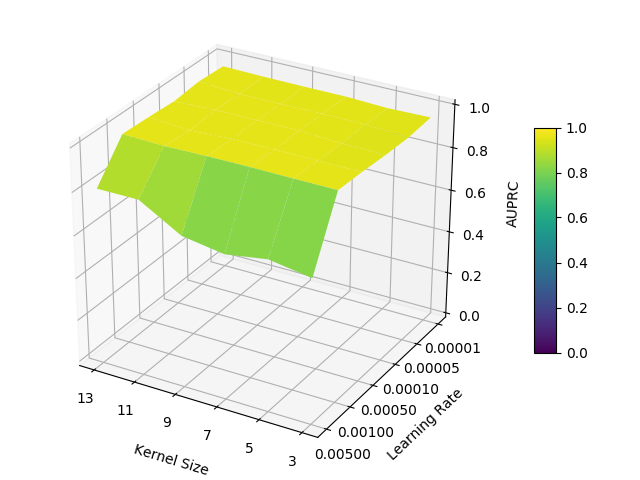} &
    			\includegraphics[width=0.30\linewidth]{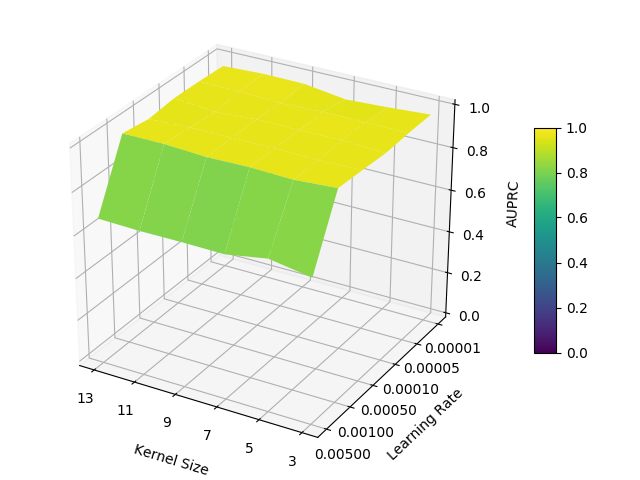} &
    			\includegraphics[width=0.30\linewidth]{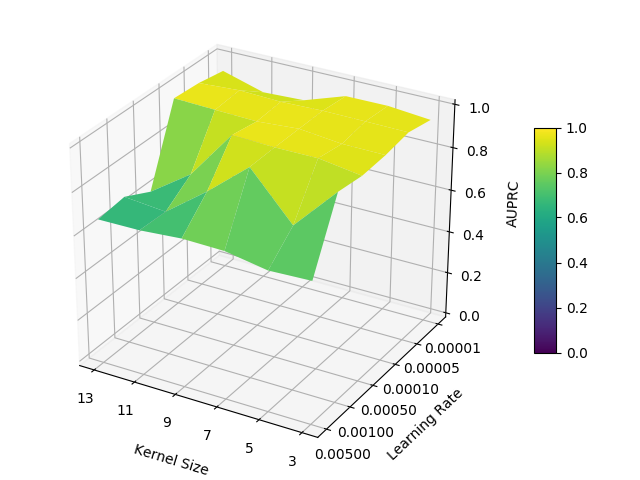} \\
			\midrule
    			\includegraphics[width=0.30\linewidth]{figures/surface-prc-ganomaly-lay0-sk} &
    			\includegraphics[width=0.30\linewidth]{figures/surface-prc-ganomaly-lay1-sk} &
    			\includegraphics[width=0.30\linewidth]{figures/surface-prc-ganomaly-lay2-sk} \\
			\midrule
    			\includegraphics[width=0.30\linewidth]{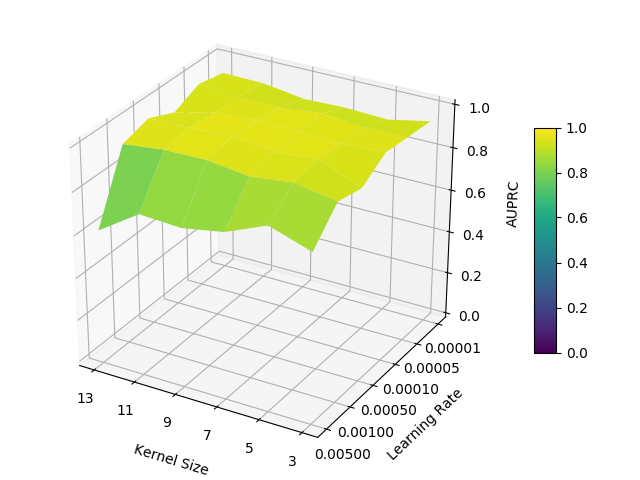} &
    			\includegraphics[width=0.30\linewidth]{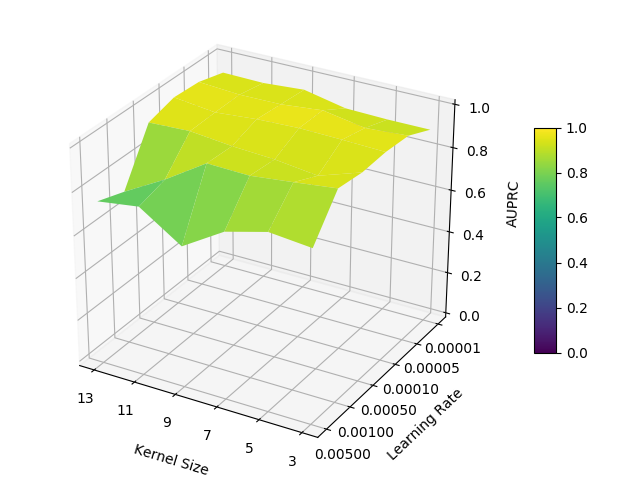} &
    			\includegraphics[width=0.30\linewidth]{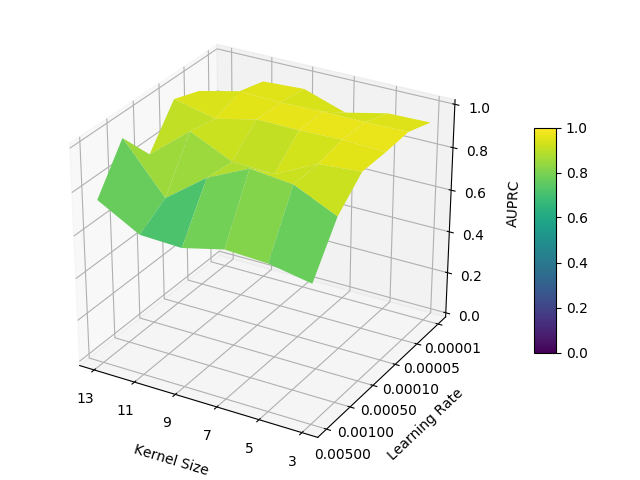} \\
		\end{tabular}
	\end{center}
	\vspace*{-5mm}
	\caption{The surface of AUPRC with various hyperparameters. The order of the shown contents is the same as Figure~\ref{fig:hyperparameter_auroc_sk}.}
	\label{fig:hyperparameter_auprc_sk}
\end{figure}

We present the measured performance for each hyperparameter with surface form in Figure~\ref{fig:hyperparameter_auroc_sk} and \ref{fig:hyperparameter_auprc_sk}. The height of the surface represents the anomaly detection ability; the higher height means higher performance. The flatten surface indicates that the neural network stably responds to hyperparameter changes.

When the number of the convolutional block is two or three, the surfaces for each architecture look like similar and relatively flatten then four-block cases. However, HP-GAN shows the most high and flatten surface in the four convolutional block case. 

We also confirm that HP-GAN shows generally flatten surface in the AUPRC surfaces. Thus, we can conclude that HP-GAN may perform stably for anomaly detection than other architectures. However, we additionally conduct the experiment for confirming the average performance with the best hyperparameter of each model. The best hyperparameter set summarized in Table~\ref{table:best_hyperparameter}.

\begin{table}[h]
    \centering
    \small
    \caption{The best hyperparameter set selected from AUROC and AUPRC surface.}
    \begin{center}
		\begin{tabular}{lccc}
		    \toprule
		        \textbf{Architecture} & \textbf{Kernel size} & \textbf{\# of conv-block} & \textbf{Learning rate} \\ 
			\midrule
    			VAE & 7 & 3 (6 convolution) & 5e-4 \\
    			GANomaly & 9 & 4 (8 convolution) & 5e-4 \\
    			HP-GAN & 7 & 3 (6 convolution) & 1e-5 \\
    			VBMH-GAN & 7 & 3 (6 convolution) & 5e-5 \\
            \bottomrule
		\end{tabular}
	\end{center}
	\label{table:best_hyperparameter}
\end{table}

We repeat the experiment 30 times with the randomly shuffled dataset for Monte Carlo estimation \cite{kroese2014monte}. The performances are summarized with a mean $\pm$ standard deviation as shown in Table~\ref{table:monte_calro_sk}.

\begin{table}[h]
    \centering
    \small
    \caption{The measured AUROC, AUPRC, and MSE are provided with a mean $\pm$ standard deviation form. The experiment is conducted with the best hyperparameter set for each architecture.}
    \begin{center}
		\begin{tabular}{lccc}
		    \toprule
		        \textbf{Architecture} & \textbf{AUROC} & \textbf{AUPRC} & \textbf{MSE} \\ 
			\midrule
    			VAE & 0.918 $\pm$ 0.033 & 0.942 $\pm$ 0.028 & 0.025 $\pm$ 0.009 \\ 
                GANomaly & 0.911 $\pm$ 0.091 & 0.937 $\pm$ 0.062 & 0.027 $\pm$ 0.049 \\
                HP-GAN & \textbf{0.948 $\pm$ 0.010} & \textbf{0.967 $\pm$ 0.008} & \textbf{0.023 $\pm$ 0.016} \\ 
                VBMH-GAN & 0.944 $\pm$ 0.021 & 0.959 $\pm$ 0.020 & 0.033 $\pm$ 0.057 \\
            \bottomrule
		\end{tabular}
	\end{center}
	\label{table:monte_calro_sk}
\end{table}

In Table~\ref{table:monte_calro_sk}, the higher the mean represents better performance. On the other hand, the lower standard deviation means higher stability. The HP-GAN that we proposed in this paper shows the higher performance at every indicator. Moreover, HP-GAN has higher stability (lower standard deviation) at every indicator.

\section{Conclusion}
\label{sec:conclusion}

We experimentally show the cutting-edge performance at anomaly detection of our HP-GAN in the TEOM-based PM sensor. The HP-GAN is trained by latent vector matching with multiple hypothesis based on WTA theory. Our neural network, HP-GAN, can generate the output more clearly and consistently with avoiding blurring effect than other VB-based models when the input data is in the normal category. The mean of AUROC and AUPRC of HP-GAN are 0.037 0.080 higher than the second performance model VBMH-GAN. Also, the mean of MSE is best (lowest) among the whole architectures. Thus, we finally conclude our HP-GAN as a cutting-edge architecture for anomaly detection.

\clearpage

\clearpage

\funding{This research received no external funding.}

\acknowledgments{Thanks to all the members of our team and Planet Co., Ltd. They have supported this research via not only data collection but also equipment for the experiment.}

\conflictsofinterest{The authors declare no conflict of interest.}

\bibliographystyle{unsrt}  
\bibliography{references}






\end{document}